\documentclass{article}
\usepackage{spconf,amsmath,graphicx, amssymb,tabularx}
\usepackage{multirow}
\usepackage{enumitem}
\usepackage{cite}
\usepackage[ruled, linesnumbered]{algorithm2e}
\usepackage[small,compact]{titlesec} 
\usepackage{times}
\usepackage[small,it]{caption}

\usepackage{comment}

\newcolumntype{D}{>{\centering\arraybackslash}m{6ex}}
\title{Transformer in action: a comparative study of transformer-based acoustic models for large scale speech recognition applications}
\name{
\begin{tabular}{c}
Yongqiang Wang$^{*}$, Yangyang Shi$^{*}$, Frank Zhang$^{*}$, Chunyang Wu$^{*}$, \\
Julian Chan$^{*}$, Ching-Feng Yeh, Alex Xiao
\end{tabular}
  \thanks{$\star$ Equal contribution.}
}
\address{Facebook AI, USA}
\begin{document}
\ninept
\maketitle
\begin{abstract}\vspace{0.5em}
In this paper, we summarize the application of transformer and its streamable variant, Emformer based acoustic model for large scale speech recognition applications. We compare the transformer based acoustic models with their LSTM counterparts on industrial scale tasks. Specifically, we compare Emformer with latency-controlled BLSTM (LCBLSTM) on medium latency tasks and LSTM on low latency tasks. On a low latency voice assistant task, Emformer gets $24\%$ to $26\%$ relative word error rate reductions (WERRs). For medium latency scenarios, comparing with LCBLSTM with similar model size and latency, Emformer gets significant WERR across four languages in video captioning datasets with 2-3 times inference real-time factors reduction.
\end{abstract}
\section{Introduction}
\label{sec:intro}

Since the introduction of deep learning into automatic speech recognition (ASR)
\cite{hinton2012deep}, architectures of the neural acoustic models have evolved from feed-forward networks (e.g., \cite{seide2011conversational}) to recurrent neural networks (e.g. long short term memory network , LSTM \cite{sak2014long}), convolution neural networks (e.g.\cite{abdel2014convolutional}) and their variants, e.g. latency-constrained bidirectional LSTM (LCBLSTM)\cite{zhang2016highway},  time-delay neural networks (TDNN) \cite{peddinti2015time}, feed-forward sequential memory networks (FSMN) \cite{zhang2015feedforward} to name a few. Recently, inspired by the great success achieved by transformer networks \cite{vaswani2017attention} in natural language processing \cite{devlin2018bert} and machine translation, self-attention based transformer acoustic models have demonstrated promising results in terms of both modeling accuracy and computational efficiency \cite{karita2019comparative, wang2020transformer, gulati2020conformer, zhang2020transformer, moritz2020streaming}. Understandably, to prove the concept, these works focused on improving the absolute word error rates (WERs) on relatively small to medium-sized public datasets, with no production constraints, such as latency and real-time factors (RTFs). This work aims to close the gap by presenting a comprehensive comparative study of transformer-based and the widely used LSTM and its variant LCBLSTM-based acoustic models on several of our internal industrial scale datasets, whose training data ranges from 9,000 (supervised) hours to 2.2 million (semi-supervised) hours. 

There is a board spectrum of speech applications powered by ASR technology,
resulting in very different requirements for ASR systems. For example, audio indexing and caption generation for video on-demand impose little to no latency constraints, but the system must be able to process a large amount of incoming traffic at a high throughput; on the other hand, users are usually sensitive to the response latency when interacting with a voice interface. As the first step in the response generation pipeline, a low latency induced by ASR systems is critical.  Caption generation for live video sits between these two extremes, where a medium latency (e.g., up to 1s) is acceptable. Different neural acoustic models have been invented to fit these applications. For example, LSTM-based neural encoders are widely used for voice interface applications due to its very small algorithmic latency; LCBLSTM-based acoustic models are widely used for medium latency applications, while BLSTM based acoustic model can be used for full-offline batch processing. 

In \cite{wang2020transformer}, we have demonstrated that the transformer-based acoustic model can significantly outperform the BLSTM-based model in terms of both modeling accuracy and inference speed. However, it is not able to process streaming audio.  In \cite{wu2020streaming}, an augmented memory transformer (AM-TRF) model is proposed to make the transformer streamable. On top of AM-TRF, \cite{shi2020mem} proposed an efficient memory transformer, Emformer, which further improves both training and inference speed. 
In this work, we will demonstrate that Emformer achieves significantly better recognition accuracies and faster RTFs on several large scale speech recognition tasks,  covering most of the low-to-medium latency scenarios. Our comparison is rigorously performed under practical production settings. We hope that by presenting these results, we can convince readers that our proposed Emformer-based acoustic models can be a competitive solution for either traditional hybrid or end-to-end style neural transducer\cite{graves2012sequence} ASR systems.

\section{Acoustic Models}
\label{sec:acousticmodel}
In both hybrid \cite{bourlard2012connectionist} and neural transducer architecture, an \emph{acoustic model}\footnote{It is also referred to as \textit{encoder} in the end-to-end modelling. We will use \textit{acoustic model} and \textit{encoder} interchangeably.} is an important component. It is used to encode an input sequence $\boldsymbol{x}_1, \cdots, \boldsymbol{x}_T$ to a sequence of high level acoustic embedding vectors $\boldsymbol{z}_1, \cdots, \boldsymbol{z}_T$. In the hybrid architecture,  these embedding vectors are used to produce a posterior distribution of tied states of hidden Markov models (HMMs), such as senone \cite{hwang1992subphonetic} or chenone \cite{le2019senones}. Cross entropy or connectionist temporal classification (CTC) \cite{graves2006connectionist} criterion can be used as the loss function. This is usually followed by a state-level minimum Bayes risk (sMBR) training \cite{kingsbury2009lattice}. In the neural transducer architecture, these acoustic embeddings are combined with the embeddings from transcription networks via a joiner network to produce a posterior distribution over the modeling units, usually characters or wordpieces \cite{schuster2012japanese}. A beam search is then performed to find the best hypothesis.  

\subsection{LSTM-based acoustic models}
In practise, unidirectional LSTM-based acoustic models are widely used in low latency ASR scenarios. Given an input sequence to the $n$-th LSTM layer, $\boldsymbol x_0^n, \cdots, \boldsymbol x_{T-1}^n$, its output (or input to the next layer) is calculated by: 
\begin{align}
    \boldsymbol x_0^{n+1}, \cdots, \boldsymbol x_{T-1}^{n+1} = \mathrm{LSTM}(\boldsymbol x_0^n, \cdots, \boldsymbol x_{T-1}^{n}; \boldsymbol c_0^n, \boldsymbol h_0^n)\label{eqn:lstm}
\end{align}
where $\mathrm{LSTM}(\cdot)$ is the function to denote LSTM cell operations on the input sequence; $\boldsymbol c_0^n$ and $\boldsymbol h_0^n$ are the initial cell and hidden state. Note that in Eq. (\ref{eqn:lstm}), $\boldsymbol x_t^{n+1}$ is a function of $\boldsymbol x_0^n$ until $\boldsymbol x_t^{n}$, thus introduce no algorithmic latency. 
In the application scenarios which allows medium latency, latency controlled bi-directional LSTM is often used. The input sequence is first chunked into a sequence of non-overlapping segments, $\mathbf C_0^0, \cdots, \mathbf C_k^0,\cdots, \mathbf C_{K-1}^0$, where $\mathbf C_k^0 \in \mathbb{R}^{c\times d}$, $c$ is the number of frames in this segment, $d$ is the dimension of individual vectors, the subscript $k$ means the index of the segments, and superscript 0 here means that these are the input the 0-th layer. To allow the acoustic model to look ahead, a right context block $\mathbf R_k^0 \in \mathbb R^{r \times d}$ with $r$ look-ahead frames is concatenated with $\mathbf C_k^0$. For the $k$-th segment, the $n$-th layer processes its input $[\mathbf C_k^n, \mathbf R_k^n]$ in the following way: 
\begin{align}
    [\overrightarrow{\mathbf C}_k^{n+1}, \overrightarrow{\mathbf R}_k^{n+1}] &= \mathrm{LSTM}_{\rm l}(
        [{\mathbf C_k^{n}}, 
        {\mathbf R_k^{n}}]; 
        \mathbf h_{k-1}^n, 
        \mathbf c_{k-1}^n
    ) \\
    [\overleftarrow{\mathbf C}_k^{n+1}, \overleftarrow{\mathbf R}_k^{n+1}] &= \mathcal{R}\left( 
        \mathrm{LSTM}_{\rm r}( 
        \mathcal{R}(
        [{\mathbf C_{k1}^{n}}, 
        {\mathbf R_{k}^{n}}]); 
        \mathbf 0, 
        \mathbf 0
    )\right) \\
    \mathbf C_k^{n+1 } = 
    [
        {\overrightarrow{\mathbf C}_k^{n+1}}^{\mathsf T},
        & {\overleftarrow{\mathbf C}_k^{n+1}}^{\mathsf T}
    ]^{\mathsf T}
     \quad  
    \mathbf R_k^{n+1 } = 
    [
        {\overrightarrow{\mathbf R}_k^{n+1}}^{\mathsf T},
        {\overleftarrow{\mathbf R}_k^{n+1}}^{\mathsf T}
    ]^{\mathsf T}
\end{align}
where $\mathrm{LSTM}_{\rm l}$ and $\mathrm{LSTM}_{\rm r}$ is the  left-to-right LSTM and right-to-left LSTM respectively. $\mathcal{R}(\cdot)$ is to reverse the order of a sequence; and $\mathbf h_{k-1}^n$ and $\mathbf c_{k-1}^n$ are the $n$-th layer's hidden and cell state at the last frame of $(k-1)$-th segment. At the last layer, $\mathbf C_0^{N-1}, \cdots, \mathbf C_{K-1}^{N-1}$ are assembled to form the sequence of acoustic embeddings.

\subsection{Transformer-based acoustic models}
For a particular transformer layer, there are two components: the self attention module and the feed-forward network (FFN) module. Given the input sequence to the $n$-th transformer layer, $\mathbf X^n = [\boldsymbol{x}_0^n, \cdots, \boldsymbol{x}_{T-1}^n]$, the self attention module performs the following operations: 
\begin{align}
    \hat{\mathbf X}^n &= \mathrm{LayerNorm}(\mathbf X^n) \\
    \mathbf Q^n &= \mathbf{W}_{\rm q} \hat{\mathbf X}^n, \; 
    \mathbf K^n = \mathbf{W}_{\rm k} \hat{\mathbf X}^n, \;
    \mathbf V^n = \mathbf{W}_{\rm v} \hat{\mathbf X}^n \\
    \mathbf Z^n &= \mathrm{Attn}(\mathbf Q^n, \mathbf K^n, \mathbf V^n) + \mathbf X^n \label{eqn:attn_ln}
\end{align}
where $\mathrm{LayerNorm}$ is the layer normalization operation~\cite{lei2016layer}, $\mathbf{W}_{\rm q}, \mathbf{W}_{\rm k}$ and $\mathbf{W}_{\rm v}$
are the projection matrices for query, key and value, respectively. $\mathrm{Attn}(\cdot)$ is the attention mechanism first proposed in 
\cite{bahdanau2014neural}. Note that Eq. (\ref{eqn:attn_ln}) uses a residual connection.  After this attention module, $\mathbf Z^n$ is fed to the FFN module equipped with layer normalization and residual connection; usually a two-layer linear transformer with Relu non-linearity is used, i.e., 
\begin{align}
    \hat{\mathbf Z}^n &= \mathrm{FFN}(\mathrm{LayerNorm}(\mathbf Z^n)) + \mathbf Z^n  \\
    \mathbf X^{n+1} &= \mathrm{LayerNorm}(\hat{\mathbf Z}^n) \label{eqn:attn_ln2}
\end{align}
Note that the last layer normalization in Eq. (\ref{eqn:attn_ln2}) is needed to prevent a path which can bypass all the transformer layers. 

The attention operation in Eq. (\ref{eqn:attn_ln}) requires access to every element in the input sequence, which makes it impossible for streaming inputs. It also incurs a quadratically growing computational cost with respect to the input sequence. To make it possible to process audio signal incrementally, \cite{wu2020streaming} proposed the augmented memory transformer (AM-TRF). Similar to LCBLSTM, AM-TRF uses block processing to deal with incrementally arrived sequences: a fixed size left and right context block, $\mathbf L_k^n \in \mathbb{R}^{l\times d}$ and $\mathbf R_k^n \in \mathbb{R}^{r \times d}$ are concatenated with the $k$-th segmeg $\mathbf C_k^n \in \mathbb R^{c \times d}$ to form a contextual segment $\mathbf{X}_k^n = [\mathbf {L}_k^n, \mathbf {C}_k^n, \mathbf R_k^n]$.
At the $k$-th segment, the $n$-th AM-TRF layer accepts $\mathbf X_k^n$ and a bank of memory vectors $\mathbf{M}_k^n = [\mathbf{m}_0^n, \cdots, \mathbf{m}_{k-1}^n]$ as the input, and produces $\mathbf X_{k}^{n+1}$ and $\mathbf {m}_k^n$ as the output, whereas $\mathbf X_{k}^{n+1}$ is fed to the next layer and $\boldsymbol m_k^n$ is inserted into the memory bank to generate $\mathbf M_{k+1}^n$ and carried over to the next segment. After all the AM-TRF layers, the center blocks $\{\mathbf {C}_k^{N-1}\}_{k=0}^{K-1}$ are concatenated as the encoder output sequence.
\begin{figure}[tbh]
    \centering
    \includegraphics[scale=0.4]{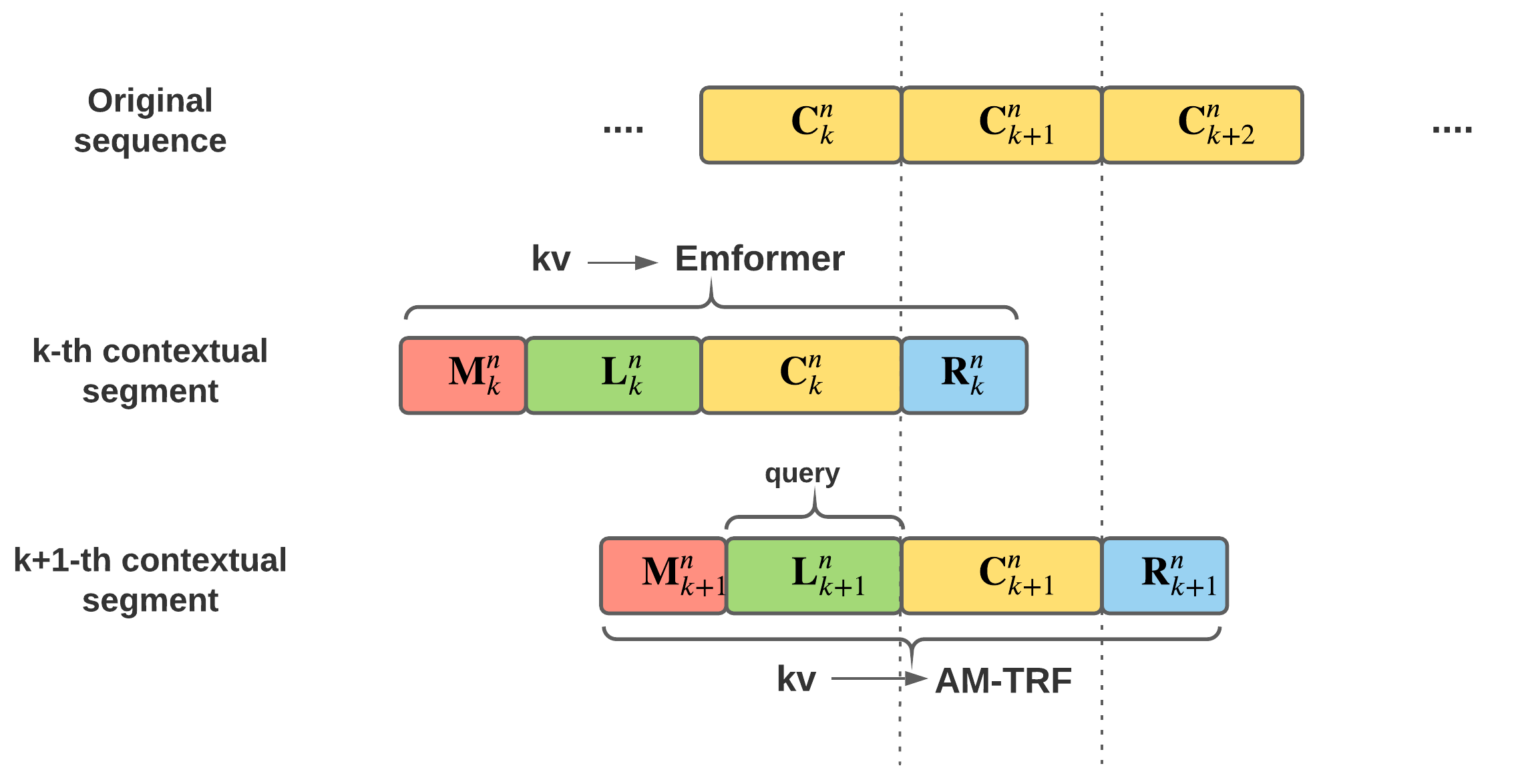}
    \caption{Comparison of AM-TRF with Emformer on computing embedding for $\mathbf L_{k+1}^n$. Emformer allows attention to previous contextual segment for query $\mathbf L_{k+1}^n$, therefore enables caching the embedding for left context blocks. }
    \label{fig:compare}
\end{figure}
\vspace{-0.5em}

One limitation of AM-TRF is that the embedding for the left context block $\mathbf L_k^{n}$ needs to be re-computed for every step, even though $\mathbf L_{k}^n$ is overlapped with $\mathbf C_{k-1}^n$ (or possibly even more previous center blocks). This is illustrated in Figure \ref{fig:compare}. In \cite{shi2020mem}, we proposed an efficient memory transformer, \textit{Emformer}, in which when compute the attention output for $\mathbf L_{k+1}^n$, the attention attend to $[\mathbf X_{k}^n, \mathbf m_k^n]$, which has been computed already in the past, therefore we can save the computation for the left context blocks, which improves both the training and inference efficiency. 
The forward pass procedure in Emformer is illustrated in Algorithm \ref{alg:emformer}. More detail of Emformer can be found in our companion paper \cite{shi2020mem}.

\vspace{-1em}
\begin{algorithm}[htb]
\For{$k$ in $0 \cdots K-1$}{
    $\bullet$ wait until $\mathbf R_k^0$ becomes available\;
    \For {$n$ in $0 \cdots N-1$}{
        $\bullet$ Apply {\rm LayerNorm} on $[\mathbf C_k^n, \mathbf R_k^n]$ to get $[\hat{\mathbf C}_k^n, \hat{\mathbf R}_k^n]$\;
        $\bullet$ Perform attention:
        \begin{align*}
            \mathbf K_k^n &= [\mathbf L_{\mathrm{k}, k}^n, \mathbf W_{\rm k} \hat{\mathbf C}_k^n, \mathbf W_{\rm k} \hat{\mathbf R}_k^n, \mathbf{W}_{\rm k}\mathbf M_k^n] \\  
            \mathbf V_k^n &= [\mathbf L_{\mathrm{v}, k}^n, \mathbf W_{\rm v} \hat{\mathbf C}_k^n, \mathbf W_{\rm v} \hat{\mathbf R}_k^n, \mathbf{W}_{\rm v}\mathbf M_k^n]
            \\
            \mathbf Z_k^n &= \mathrm{Attn}(\mathbf W_{\rm q} \hat{\mathbf X}_k^n; \mathbf K_k^n , \mathbf V_k^n ) + \mathbf X_k^n \\
            \mathbf m_{k}^{n+1} &= \mathrm{Attn}(\mathbf W_{\rm q} \mathbf s_k^n ; \mathbf K_k^{n\prime}, \mathbf V_k^{n\prime})
         \end{align*}
         \\
        $\bullet$ Update the cache 
        $\mathbf L_{\mathrm{k}, k}^n$
        to $\mathbf L_{\mathrm{k}, k+1}^n$:
        \[
            \mathbf L_{\mathrm{k}, k+1}^n = \mathrm{cancat}(\mathbf L_{\mathrm{k}, k}^n, \mathbf K_k^n)[:-l, :]
        \]
        and similarly for $\mathbf L_{\mathrm{v}, k}^n$.\\
        $\bullet$ Forward $\mathbf{Z}_k^n$ via FFN to get $\mathbf X_{k}^{n+1}$
    }
}
\caption{Forward pass of Emformer. ${\mathbf s}_{k}^{n}$ is the average pooling over $\mathbf C_k^n$; $\mathbf K_k^{n\prime}$ is obtained by excluding $\mathbf W_k^n \mathbf M_k^n$ from $\mathbf K_k^n$ and similarly for $\mathbf V_{k}^{n\prime}$.}
\label{alg:emformer}
\end{algorithm}
\vspace{-1em}

\section{Experiments}
\label{sec:exp}

This section compares the LSTM-based acoustic models with the transformer-based acoustic models for both low-latency and medium latency scenarios. Measuring user perceived latency is not straightforward and can be impacted by many aspects of the ASR systems. In this work, we define encoder induced latency (EIL) as the average difference between the time we received the physical signals and the time when the encoder emits the corresponding embeddings, assuming the actual computation is infinitely fast. In this way, we can focus on comparing the algorithmic latency introduced in different encoders. Under this definition, the EIL of models using block processing is calculated as the size of the right context plus half of the length of center segment.  We compare word error rates (WERs) and real-time factors (RTFs) under the fixed EIL. RTFs are measured on a host with Intel Xeon D-2191A 18-core CPUs, while only 2 CPU threads are used for one utterance, one for forwarding through the acoustic model, and the other for search; 10 utterances are concurrently decoded when measuring RTFs. 

All the experiments are using 80-dimensional log-Mel filter bank features at a 10ms frame rate. Speed perturbation \cite{ko2015audio} and SpecAugment \cite{park2019specaugment} without time warping are also used in all experiments.  We compare encoders using different training criteria and/or modeling units. For hybrid systems, CTC criterion is used to train either a context-dependent grapheme (a.k.a. chenone) or a context-independent wordpiece systems \cite{zhang2020fast}. For neural transducer systems, wordpieces are used as the modeling unit. Kaldi\cite{Povey_ASRU2011} recipe is used to bootstrap the chenone system, while wordpiece systems do not need a bootstrap process. All the neural network models are trained using Adam optimizer \cite{kingma2014adam}. Unless noted otherwise, all the models are trained by scanning the training data for 100 to 200 epochs. Learning rates increase to 1e-3 in 20K warming-up updates; then, it is fixed for 60 to 80 epochs; after that, the learning rate shrinks every epoch with factor 0.95. In training deep transformer models, an auxiliary incremental loss~\cite{Andros2019} with weight 0.3 is used. All the models are trained using 32 Nvidia V100 GPUs with fp16 precision, except for the English and Spanish video ASR tasks, where 64 GPUs are used. Our training throughput ranges from 4M to 8M frames per second, which allows us to finish most of the experiments in less than 3 days. 
To get the best RTFs, all trained models are quantized to INT8, where each channel (row) in a matrix has its own quantizer. Compared with the fp16 model, per-channel quantization roughly costs us 1.2\% relative WER loss but doubles the neural network evaluation speed. More details about our training and inference setup can be found in \cite{wang2020transformer, shi2020mem}.

\vspace{-0.5em}
\subsection{Comparison in Low latency Tasks}
\vspace{-0.5em}

To compare encoders in the low latency scenarios, we used our internal personal assistant dataset that contains queries like calling (e.g., “call Alex”) and other types of voice queries (e.g., “what’s the weather in Menlo Park?”). 
Our training set is made up of two manually-transcribed de-identified in-house corpora with no personally identifiable information (PII). The first corpus comprises 15.7M utterances (12.5K hours) in the voice assistant domain recorded by 20K crowd-sourced workers on mobile devices. The second corpus contains 1.2M voice commands (1K hours)
sampled from the production traffic of a smart speaker after
the wakeword is activated. Utterances from this corpus are further morphed in an effort to de-identify the user. We distort each utterance in the first
corpus twice using simulated Room Impulse Response (RIR) and additive background noise extracted from public videos. Utterances in the second corpus are already processed by a far-field frontend, so there is no need for RIR distortion. We
create a few copies of each utterance in this corpus using speed perturbation and add additive background noise. The final distorted train data contains 56.1M utterances (47K hours). The evaluation data consists of 22.3K hand-transcribed
de-identified utterances from volunteer participants in the in-house program, which consists of employee
households that have agreed to have their voice activity reviewed and tested. Every utterance in this set has
an associated contact list, which we use for on-the-fly personalization for calling queries. These evaluation utterances are divided into two parts: \texttt{name} comprises 7K utterances that have at least one entity name in the transcription, and the reset 15.3K utterances form the \texttt{general} test set. 

Uni-directional LSTM and Emformer-based hybrid systems are trained on this task using 8K chenones as the modeling units. The current frame is stacked with 7 future frames as the input feature vector to the LSTM network. After the first layer, the input sequence is sub-sampled at a rate of 4:1. The LSTM network has 5 layers, each with 1200 cells, resulting in a model size of 65M parameters. To speed up inference, 10 frames (100ms) are grouped as a batch for one forward through the LSTM network. Together with the 7-frame look-ahead in the frontend, this yields an EIL of 120ms. In Emformer, a linear layer first projects the 80-dimensional feature vectors to 128 dimensions, then every 4 such vectors are stacked and form a sequence of 512-dimensional feature vectors at a frame rate of 40ms. Two Emformer configurations which yield similar EILs as the baseline LSTM are used: in the first configuration, the center segment size is set as 120ms, and the right context is 80ms, which gives an EIL of 140ms; in the second configuration center segment size is 80ms, and the right context is 40ms, which gives an EIL of 80ms. Both configurations use a left context length of 800ms. Since the left context covers 7 to 10 past segments, there is no need to use memory in this scenario, thus we set the memory bank size to 0. Both configurations use 18 layers of Emformer layers with 8 attention heads and 64 dimensions per head and 2048 as the FFN network dimension, yielding models with 60M parameters. 

\vspace{-1em}
\begin{table}[ttbph]
    \centering
    \begin{tabular}{|c|cc||cc|c|}
    \hline
    Train & \multirow{2}{*}{Model} & EIL & \multicolumn{2}{c|}{test set} & \multirow{2}{*}{RTF}  \\
    (hrs) &     & (ms) & \texttt{name} & \texttt{general} & \\
    \hline\hline
    \multirow{3}{*}{4.7K}  & LSTM & 120 & 10.89 & 5.43 & 0.194 \\
                           & \multirow{2}{*}{ Emformer} & 140 & 8.05 & 3.94 & 0.227 \\
                           &                           & 80 & 8.21 & 4.12 & 0.276 \\
    \hline
    47K                     & Emformer & 80 & 6.83 & 4.05 & 0.275  \\
    \hline
    \end{tabular}
    \caption{WER/RTF comparison of LSTM with Emformer in a task requires low latency.}
    \label{tab:portal}
\end{table}
\vspace{-1em}

Results are presented in Table \ref{tab:portal}. In the first block, we first performed a quick turn-around experiment where only 10\% of training data is used. With similar EILs (120ms vs 140ms), Emformer model significantly outperforms the LSTM baseline by 24-26\% on both test sets, with slightly increased RTF. When further reducing the latency induced by encoder to 80ms, WERs are slightly increased. It also comes at the cost of RTFs (0.227 to 0.276) since the Emformer network needs to be evaluated more frequently on small batches. Lastly, by scaling up the training data to 47K hours, another 16.8\% WER reduction is observed on the \texttt{name} test set.

\vspace{-0.5em}
\subsection{Comparison in Medium Latency Tasks}
\vspace{-0.5em}

For medium latency scenarios, we compare the LCBLSTM-based encoder with Emformer on our internal \textit{video ASR} task. 4 languages, \textit{en} (English), \textit{es} (Spanish), \textit{vi} (Vietnamese) and \textit{de} (German), are used in the comparison. The training data are extracted from videos shared publicly by users, where only the audio part is used. These data
are de-identified: both transcribers and researchers do not
have access to any user-identifiable information. For each language, besides a \textit{dev} set for development purpose,  there are  3 test sets in different conditions: \textit{clean}, \textit{noisy} and \textit{extreme}. A voice activity detector (VAD) is used to segment whole audios into chunks which are not longer than 45 seconds\footnote{Hybrid systems with LCBLSTM and Emformer can decode unlimited length of audio with only 2-3\% relatively WER increase.}. Decoding is performed on these chunks. Training data are first aligned against reference hypothesis, and segmented into chunks with a maximum duration of 10 seconds. 
Training and test set sizes are summarized in Table \ref{tab:video_asr}. Furthermore, on \textit{en}, in additional to the 39.4K hours supervised training data, we also prepared 2.16M hours unsupervised training data, whose transcription is obtained by sending de-identified, user-uploaded public videos to our internal automatic transcription service. No human effort is involved in transcribing these unsupervised data. In total, we have 2.2M hours semi-supervised data for the \textit{en} task. To the best authors' knowledge, this is so far the largest speech recognition task ever published.   

\vspace{-1em}
\begin{table}[htbp]
    \centering
    \begin{tabular}{|c||c|ccc|c|}
    \hline
    language & Train & \textit{clean} & \textit{noisy}  & \textit{extreme}   \\
    \hline\hline
    \textit{de} & 9.1K & 24.9 & 24.7 & 42.3   \\
    \textit{vi} & 10.2K & 24.3 & 24.5 & 49.6  \\
    \textit{es} & 21.2K & 26.7 & 26.3 & 49.7  \\
     \textit{en} & 39.4K & 24.2 & 24.4 & 46.6   \\
    \hline 
    \end{tabular}
    \caption{Training data and test sets sizes (in hours) for 4 languages. }
    \label{tab:video_asr}
\end{table}
 \vspace{-1em}

Due to the differences in language characteristic, latency requirement, and size of training data, the architecture of ASR system for these four languages are slightly different: 
\begin{itemize}[wide, labelwidth=!, labelindent=0pt]
    \item For \textit{vi} sub task, chenone systems are trained using the CTC criterion. The LCBLSTM encoder consists of 5 layers, each with 800 hidden dimensions per direction. After the first and second layers, two 2:1 sub-samplings are performed, respectively. This yields a stride 4 encoder. Input to the Emformer network is first projected to 128-dimension, then 4 frames are stacked together to form an input sequence with a 40ms frame rate.  
    The Emformer encoder consists of 26 layers, each with 8 attention heads and 64 dimensions per head. During both training and inference, the maximum memory bank size is 4. For both LCBLSTM and Emformer, the center segment size is 1480ms, and the right context size is 320ms, i.e., EIL is 1060ms. The left context size in Emformer is 800ms.    
    \item For \textit{de} and \textit{es} sub tasks, we trained neural transducer systems with 2047 wordpieces following setups in \cite{zhang2020benchmarking}. The LCBLSTM and Memformer encoders architecture follows the \textit{vi} sub task, except the center segment size is 800ms, and the right context size is 320ms, which means EIL is 720ms. The left context size in Emformer is 800ms. In the predictor of the transducer, the wordpiece tokens are first represented by 256-dimensional embeddings before going through two LSTM layers, each with 512 hidden nodes,  followed by a linear projection to 1024-dimension before the joiner; in the joiner, the combined embeddings from the encoder and the predictor first go through a \emph{Tanh} activation and then another linear projection to 2048. For deeper models such as transformer and Emformer, both cross entropy loss and neural transducer loss are used to compute auxiliary loss, which is found to improve WERs\cite{liu2020improving}.

    \item For \textit{en} task, hybrid systems are trained using CTC criterion and wordpiece as the modeling unit. As demonstrated in  \cite{zhang2020fast}, a large stride 8 in encoder can be used to speed up the inference without much impact on WER. Since \textit{en} task has more training data, we increase the model capacity for both LCBLSTM and Emformer: LCBLSTM has 6 layers with each layer 1000 hidden cells; three 2:1 subsampling is performed after the first three layers respectively; Input to the Emformer network is first projected to 64-dimension, then 8 frames are stacked together to form an input sequence with an 80ms frame rate. Emformer's depth is increased to 36 layers. The center segment size and right context size are kept the same as the one in \textit{es} and \textit{de} sub tasks. 
    Furthermore, we also trained both LSTM-based and transformer-based acoustic models on the 2.2M hours semi-supervised data. Due to the sheer volume of training data, we only scan the training data by 3 epochs, and save the checkpoints for every 5K steps.
    This allows us to finish training in 5 days. 
    
    \item For \textit{vi}, \textit{es} and \textit{en}, the corresponding systems which use transformer-based encoders are also trained to show the upper bound of transformer-based acoustic models. 3 VGG blocks are used as not only a learnable frontend but also to add relative positional embedding information to the input of transformer layers. See \cite{wang2020transformer} for more details. For \textit{vi} and \textit{es}, 24 transformer layers are used. For \textit{en}, 36 transformer layers are used.  
\end{itemize}

Results are presented in Table \ref{tab:video_asr_results}. Under the same latency constraint, except for \textit{extreme} on the \textit{vi} sub task, Memformer consistently outperforms LCBLSTM, yielding 11.2\% - 19.8\% WER reductions (WERR) on \textit{clean}, 9.3\% to 18.1\% on \textit{noisy} and 3.6\% to 15.6\%  on \textit{extreme}.  Emformer also enjoys the fully parallelized computation within each forwarding step, which reduces the inference RTF over LCBLSTM by 2 to 3 times. On the other hand, compared with transformer-based  model which does not have latency constraint, 
Emformer stills consistently lags behind by 3\% to 11\%, which indicates there is still some space for accuracy improvement. Lastly, when training on the 2.2M-hour semi-supervised \textit{en} data, both LCBLSTM and Emformer get over 10\% WER reduction on \textit{extreme} test set over the supervised baselines; transformer gets even large gains, 7.4\% on \textit{clean}, 10\% on \textit{noisy} and 15\% on \textit{extreme}.  Again the increased gap between transformer and Emformer indicates there are still some room of improvement for streamable transformers. 

\begin{table}[htbp]
    \centering
    \begin{tabular}{|c|c||ccc|c|}
    \hline
    Lang/System & Encoder & \textit{clean} & \textit{noisy} & \textit{extreme} &  RTF \\
    \hline\hline
    \emph{de} &  LCBLSTM &  13.39   & 14.46 & 15.71  & 0.22 \\
    transducer    &  Emformer &   10.74 & 11.84   & 13.26  & 0.11 \\
    \hline
    \multirow{2}{*}{\emph{vi}} &  LCBLSTM &  11.34   & 16.12 & 40.96 & 0.19 \\
                              &  Emformer   &  9.69 & 14.62 & 42.36 & 0.09 \\
     hybrid         & Trf  & 9.00  & 13.71 & 39.27 & 0.09 \\
    \hline
        \multirow{2}{*}{\emph{es}} &  LCBLSTM   & 10.24 & 13.58 & 23.23  & 0.22 \\
                              &   Emformer   & 8.89  & 12.01 & 20.70 &  0.11\\
        transducer   & Trf  &  8.62  & 11.63 & 19.78 &  0.11 \\
    \hline
            \multirow{2}{*}{\emph{en}} &  LCBLSTM  & 11.90 & 16.06 &  22.49 & 0.27 \\
                              &  Emformer  &  10.56  & 14.51 & 21.69 &  0.09\\
            hybrid            & Trf  &  9.96  & 13.85 & 20.80 &  0.11 \\
    \hline
            \emph{en} &  LCBLSTM & 10.56 & 14.29 &  19.73 & 0.27 \\
            hybrid    &  Emformer  &  10.54  & 13.98 & 19.51 &  0.09\\
                             (2.2M hrs) & Trf  &  9.22  & 12.47 & 17.68 &  0.09 \\
    \hline
    
    \end{tabular}
    \caption{Comparison of LCBLSTM with Emformer in medium latency scenarios. Transformer, which can be only used in offline scenarios, are also shown in the table for \textit{vi}, \textit{es} and \textit{en}. }
    \label{tab:video_asr_results}
\end{table}

\vspace{-1.5em}

\section{Conclusions}
\label{sec:con}
\vspace{-0.5em}
In this work, we compare the LSTM-based acoustic models with transformer-based ones for a range of large scale speech recognition tasks. Our results show that  for  low  latency, voice assistant task,  Emformer, a streamable transformer, gets 24\% to 26\% relative WERRs, compared with LSTM.  For medium latency, video captioning task, compared with LCBLSTM,  Emformer  gets significant WERR across four languages and 2-3 times RTF reduction. Results on a task with 2.2M hours semi-supervised training data, indicate that there are still room for improvement for Emformer. Combining with convolution network to reduce model size and computational cost (e.g. \cite{yeh2020streaming}) can be our future work.

\bibliographystyle{IEEEbib}
\bibliography{strings,refs}

\end{document}